\documentclass[runningheads]{llncs}
\usepackage{caption}
\usepackage{subcaption}
\usepackage{graphicx}
\usepackage{amsmath}
\usepackage{multirow}
\usepackage[dvipsnames]{xcolor}

\begin{document}
\title{Revisiting FunnyBirds evaluation framework for prototypical parts networks\thanks{This work was funded by the National Science Centre (Poland) grant no. 2022/47/B/ST6/03397. We gratefully acknowledge Polish high-performance
computing infrastructure PLGrid (HPC Centers: ACK Cyfronet AGH) for providing computer facilities and support within computational grant no. PLG/2023/016555.}}
%
%
\author{Szymon~Opłatek\inst{1} \and Dawid~Rymarczyk\inst{1,2}\orcidID{0000-0002-8543-5200} \and
Bartosz~Zieliński\inst{1,3}\orcidID{0000-0002-3063-3621}}
\authorrunning{Opłatek S. et al.}
%
\institute{Jagiellonian University, Faculty of Mathematics and Computer Science, Łojasiewicza 6, 30-348, Poland \\
\email{szymon.oplatek@student.uj.edu.pl}
\email{\{dawid.rymarczyk;bartosz.zielinski\}@uj.edu.pl}
\and
Ardigen SA, Podole 76, 30-394, Kraków, Poland
\\ \and
IDEAS NCBR, Chmielna 69, 00-801 Warsaw, Poland\\}
\maketitle              
\begin{abstract}
Prototypical parts networks, such as ProtoPNet, became popular due to their potential to produce more genuine explanations than post-hoc methods. However, for a long time, this potential has been strictly theoretical, and no systematic studies have existed to support it. That changed recently with the introduction of the FunnyBirds benchmark, which includes metrics for evaluating different aspects of explanations.
However, this benchmark employs attribution maps visualization for all explanation techniques except for the ProtoPNet, for which the bounding boxes are used. This choice significantly influences the metric scores and questions the conclusions stated in FunnyBirds publication.
In this study, we comprehensively compare metric scores obtained for two types of ProtoPNet visualizations: bounding boxes and similarity maps. Our analysis indicates that employing similarity maps aligns better with the essence of ProtoPNet, as evidenced by different metric scores obtained from FunnyBirds. Therefore, we advocate using similarity maps as a visualization technique for prototypical parts networks in explainability evaluation benchmarks.

\end{abstract}

\section*{Errata}
        
After our paper was accepted at the XAI 2014 conference, we discovered an inaccuracy that needs clarification and correction. While this inaccuracy does not affect the interpretation or conclusions of our results, it is important for the community to be informed about it.

We submitted our paper to XAI 2014 before the authors of FunnyBirds framework~\cite{hesse2023funnybirds} published their full source code. That is why we had to reimplement the code of the ProtoPNet experiment using the description from the FunnyBirds paper. We also had to train a set of ProtoPNet models, as they were also not published. As a result, our results were obtained for the reimplemented experiment and our ProtoPNet models.

After submitting our paper to XAI 2014, the FunnyBrids authors published the code of the ProtoPNet experiment and the ProtoPNet model. However, their source code contained a critical error, significantly changing the metrics values. We reported it as an issue at the FunnyBirds GitHub repository\footnote{ \url{https://github.com/visinf/funnybirds/issues/5}}. After fixing this error, we achieved higher metrics values, as presented in Table~\ref{tab:erratA}.

\begin{table}[]
    \centering    
    \caption{SD and TS metrics values for the original FunnyBirds (FB) code with and without error differ significantly.}
    \begin{tabular}{llclc}
    \hline
    \textbf{Metric} & \hspace{.25cm} & \textbf{FB code with error} & \hspace{.5cm} & \multicolumn{1}{l}{\textbf{FB code without error}} \\ \hline
    Accuracy        &                            & 0.94                                        &                            & 0.94                                                           \\
    BI              &                            & 1.00                                        &                            & 1.00                                                           \\ 
    CSDC            &                            & 0.93                                        &                            & 0.93                                                           \\
    PC              &                            & 0.91                                        &                            & 0.91                                                           \\
    DC              &                            & 0.92                                        &                            & 0.93                                                           \\
    D               &                            & 0.58                                        &                            & 0.58                                                           \\ 
    SD              &                            & 0.24                                        &                            & \textbf{0.75}                                                  \\ 
    TS              &                            & 0.46                                        &                            & \textbf{0.56}                                                  \\ \hline
    \end{tabular}
    \label{tab:erratA}
\end{table}

Moreover, after incorporating our Summed Similarity Maps to FunnyBirds code without error, we obtained results presented in Table~\ref{tab:main_errata}, which confirm the conclusions presented in our XAI 2014 conference paper.

\begin{table}[]
    \centering 
    \caption{After fixing the error of FunnyBirds (FB) code and incorporating our Summed Similarity Maps, the main conclusions of our paper hold, i.e., values of D, SD, and TS rise, while scores for CDSC, PC, and DC drop. The reason for that is explained in the Subsection~\ref{abcd}. The table below should be considered instead of the Table~\ref{tab:main}.}
    \begin{tabular}{llclc}
    \hline
    \textbf{Metric} & \hspace{.5cm} & \textbf{BB (FB code w/o error)} & \hspace{.5cm} & \textbf{SSM (FB code w/o error)} \\ \hline
    Accuracy        &                            & 0.94                                &                            & 0.94                                 \\
    BI              &                            & 1.00                                &                            & 1.00                                  \\
    CSDC            &                            & \textbf{0.93}                                &                            & 0.89                                 \\
    PC              &                            & \textbf{0.91}                                &                            & 0.84                                 \\
    DC              &                            & \textbf{0.93}                                &                            & 0.89                                 \\
    D               &                            & 0.58                                &                            & \textbf{0.61}                                 \\
    SD              &                            & 0.75                                &                            & \textbf{0.83}                                 \\
    TS              &                            & 0.56                                &                            & \textbf{0.64}                                 \\ \hline
    \end{tabular}
    \label{tab:main_errata}
\end{table}

We believe these corrections will lead to a more accurate understanding of the ProtoPNet model as well as the FunnyBirds evaluation framework. The rest of the work is as it was originally published at the XAI 2014 conference.

\section{Introduction}

Standard deep neural networks (DNNs) lack transparency in their decision-making process, posing challenges for human verification, especially in critical domains such as medicine~\cite{komorowski2023towards,rudin2019stop}. In response, the field of eXplainable Artificial Intelligence (XAI) has emerged with post-hoc and ante-hoc methods. Post-hoc methods are commonly used because they can be applied to already-trained neural networks. However, various studies have highlighted their potential biases~\cite{adebayo2018sanity,arras2022clevr,tomsett2020sanity}, raising concerns about the reliability of their explanations. Consequently, ante-hoc methods like ProtoPNet~\cite{chen2019looks} and B-Cos~\cite{bohle2022b} have gained prominence.

These intrinsically interpretable or self-explainable methods operate under the assumption that the model's design inherently allows for interpretable predictions. However, they often require more complex training to achieve comparable accuracy to standard DNNs, leading to the interpretability-accuracy trade-off phenomenon~\cite{rudin2022interpretable}.

\begin{figure}[t]
    \centering
    \includegraphics[width=0.85\textwidth]{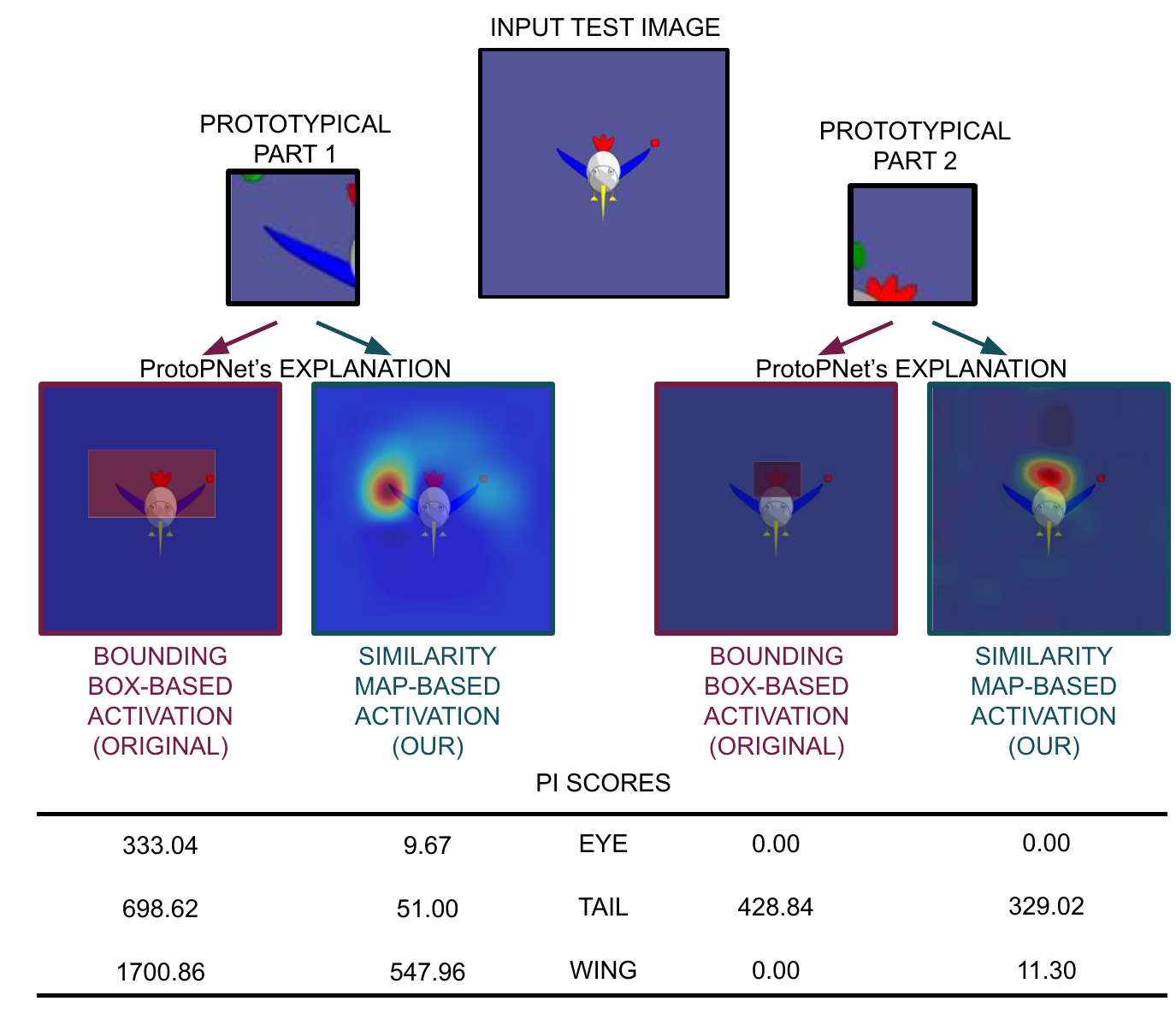}
    \caption{Attribution Maps (AM) based on bounding boxes and similarity maps for two prototypical parts of ProtoPNet trained on the FunnyBirds dataset. For prototypical part 2, both AM types correctly cover the tail prototype. However, for prototypical part 1, AM based on bounding boxes incorrectly covers almost the whole area of the bird. Such discrepancy results in incorrect values of interface function PI (e.g. $333.04$ instead of $0$ for eyes) and inaccurate values of FunnyBirds metrics (see Section~\ref{sec:method}).}
    \label{fig:teaser}
\end{figure}

Practitioners encounter a dilemma regarding whether to choose a standard DNN coupled with a post-hoc explanation method to achieve higher accuracy or to invest in the development of a self-explanatory model to enhance interpretability. Addressing this question necessitates a reliable and trustworthy evaluation framework for model explanations. This challenge is tackled by the FunnyBirds framework~\cite{hesse2023funnybirds} that introduces a synthetic dataset and a set of metrics for comparing explanation quality across different models, including post-hoc and interpretable ones.

However, a limitation of the FunnyBirds evaluation lies in how metrics are computed for the ProtoPNet model compared to other explanation methods such as GradCAM~\cite{selvaraju2017grad} and LRP~\cite{bach2015pixel}. The assumption made by the authors is that ProtoPNet explanations are presented as bounding boxes highlighting important image regions. However, these bounding boxes only approximate the significance of regions derived from more precise similarity maps, as illustrated in Figure~\ref{fig:teaser}, which can be seen as equivalent to saliency maps for post-hoc methods.

In this study, we evaluate ProtoPNet explanations based on similarity maps rather than bounding boxes within the FunnyBirds framework and comprehensively analyze the resulting changes in explanation quality. Our findings demonstrate that similarity map-based explanations better align the metrics with ProtoPNet's design intuition, yielding more accurate evaluation results. Therefore, we advocate for adopting similarity map-based activations for ProtoPNet evaluations to ensure a reliable comparison of explanations within the community.

\section{Related works}

\paragraph{Evaluation of xAI.}
With the advancements in xAI methodologies, the need to quantify the quality of provided explanations has emerged. Benchmarking xAI approaches can be categorized into two main groups: those based on user studies and those involving the development of dedicated quantitative metrics.

Evaluation through user studies has been explored in previous research, e.g. in~\cite{kim2018interpretability}, the correctness of explanations was assessed, while~\cite{jeyakumar2020can} delved into determining the most suitable form of explanation for different data types. Additionally, in~\cite{kim2022hive}, the level of user overconfidence induced by explanations was measured. More specific evaluations include assessing semantic similarity for prototypical parts in~\cite{rymarczyk2021protopshare}, examining explanation saliency in~\cite{rymarczyk2022interpretable}, and evaluating the adequateness of prototypical parts for the medical domain in~\cite{nauta2023interpreting}.

On the other hand, in proposing metrics and taxonomies for evaluating explanations, the Co-12 framework was introduced in~\cite{le2023benchmarking,nauta2023anecdotal,nauta2023co}. This framework provides a taxonomy for explanation evaluation and analyzes existing approaches such as Quantus~\cite{hedstrom2023quantus}, Ablation~\cite{hameed2022basedxai}, and OpenXAI~\cite{agarwal2023evaluating}. While these approaches predominantly focus on general toolkits for assessing explanation quality across multiple models and data modalities, there are also works proposing metrics specifically designed for prototypical parts, such as purity \cite{nauta2023pip} and spatial misalignment \cite{sacha2023interpretability}.

However, recent work such as~\cite{hesse2023funnybirds} aims to compare explanations among different methods using synthetic datasets. Nonetheless, the assumptions made for prototypical parts in this framework do not entirely align with the ProtoPNet essence. Thus, we propose a different method to derive them to ensure fair comparability with attribution-based methods.

\section{Methods}
\label{sec:method}

\subsection{FunnyBirds}

\paragraph{Dataset.}
FunnyBirds dataset consists of synthetically generated bird images rendered from five human comprehensible concepts of beak, wings, feet, eyes, and tail, called parts. The dataset contains 50 bird classes, each corresponding to a unique subset of 26 predefined parts. In total, it comprises 50,000 training images and 5,000 testing images in $256\times 256$ resolution. Furthermore, the training set incorporates augmented images with missing bird parts, simulating a data mix-up strategy.

\paragraph{Interface functions.}
The second major aspect of the framework are interface functions, $PI(\cdot)$ and $P(\cdot)$. These functions are designed to translate various explanation types (such as saliency maps or prototypical parts) into a unified format that can be used to calculate explainability metrics. Based on an explanation, the $PI(\cdot)$ function calculates a set of importance scores assigned to each part, while the $P(\cdot)$ function provides a set of important parts parameterized by the threshold $t$ used to control the ``sensitivity'' of importance.

\paragraph{Default interface functions for prototypical parts.}
FunnyBirds authors introduce the default definition of interface functions for specific XAI methods. For prototypical part-based methods, they calculate $PI(\cdot)$ by summing the values of an attribution map within particular bird parts. The attribution map is obtained as follows for a training sample $(x,y)$: the image $x\in X$ is passed to ProtoPNet; for each prototypical part corresponding to class $y$, we obtain a similarity map and corresponding bounding box; such a bounding box is then filled with the maximum value multiplied by the weight between the prototypical part and class $y$; the attribution map is obtained as a sum of such bounding boxes obtained for all prototypical parts. When it comes to $P(\cdot)$, it is defined as a set of bird parts with at least $t$-percent of area overlapping the union of those bounding boxes.

\begin{figure}[t]
    \centering
    \includegraphics[width=0.9\textwidth]{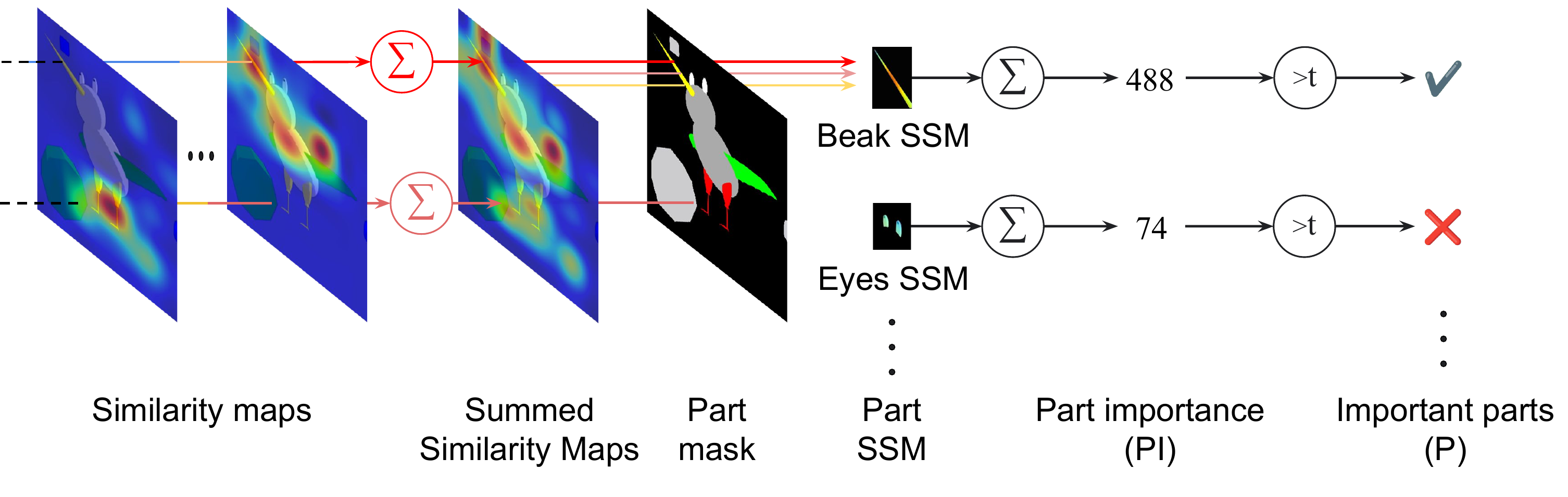}
    \caption{Calculating Summed Similarity Maps (SSM) and interface functions $PI(\cdot)$ and $P(\cdot)$. The process starts with generating SSM by summarizing the similarity maps obtained for prototypical parts. Then, for each bird part, we multiply its mask with SMM and sum it up to obtain part importance (e.g. part importance for beak equals 488). To obtain important parts $P(\cdot)$, we analyze which of them has importance higher than the considered threshold $t$ (e.g., eyes are not in $P(\cdot)$ because their importance 74 is smaller than the threshold). For this example, $PI=\{beak: 488, eyes: 74, legs: 371, \dots\}$ and $P=\{beak, legs, wings\}$.}
    \label{fig:Attrib_map}
\end{figure}

\paragraph{Metrics.}
The FunnyBirds metrics design follows the Co-12 taxonomy~\cite{nauta2023co}, exploring categories such as Completeness, Correctness, and Contrastivity. The latter two are measured by Single Deletion (SD) and Target Sensitivity (TS) metrics, respectively. At the same time, the Completeness score is calculated as an average of Controlled Synthetic Data Check (CSDC), Preservation Check (PC), Deletion Check (DC), and Distractibility (D).
Here, we recall the definition of one of the metrics, namely SD, to build an intuition on how the $PI(\cdot)$ and $P(\cdot)$ are used:
\begin{equation}
\label{eq:sd}
\text{SD} = \frac{1}{2} + \frac{1}{2|X|}\sum_{x\in X} \rho(\text{PI}(e), f(x) - \{f(x_{\setminus p})\}_{p}),
\end{equation}
where $e$ denotes the explanation received for image $x$, $f(x)$ is the logit of class $y$, and $f(x_{\setminus p})$ is the same logit obtained after removing part $p$ of the bird. Finally, the $\rho$ is the Spearman rank-order correlation between two sorted sets.

\subsection{Summed Similarity Maps (SSM) for more precise interface functions}

We propose an alternative definition of the interface functions based on the similarity maps, which are more precise than bounding boxes, as presented in Figure~\ref{fig:teaser}. Similarly, like in the default definition, $PI(\cdot)$ is calculated by summing the values of an attribution map within particular bird parts. However, our definition of attribution map differs as follows: the image $x\in X$ is passed to ProtoPNet; for each prototypical part corresponding to class $y$, we obtain a similarity map; such similarity map is then multiplied by the weight between the prototypical part and class $y$; the attribution map is obtained as a sum of such similarity maps obtained for all prototypical parts. We call this approach Summed Similarity Maps (SSM).

Regarding $P(\cdot)$, we decided to reuse SSM. Therefore, for a given threshold $t$, a part is considered important if the sum of SSM pixels overlapping this part is larger than $t$-percentage of a total SSM sum. The calculation process is presented in Figure~\ref{fig:Attrib_map}.

\section{Experimental setup}

We use the ProtoPNet model~\cite{chen2019looks} with ResNet50, VGG19, and DenseNet169 backbones. We follow the training setup from FunnyBirds framework~\cite{hesse2023funnybirds}. It corresponds to the multilabel classification because input images present incomplete birds fitting more than one class.
We use Adam optimizer~\cite{kingma2017adam} with a learning rate decreasing every 10th epoch, and we apply prototype projection at the 25th epoch. Moreover, it is trained three times with different prototype sizes (128, 256, or 512) but with the same number of prototypical parts equal to 10. We do not use any augmentations.

The code is publicly available\footnote{ \url{https://github.com/hamer101/FunnyBirds\_PrototypesRevisited}}. The training was conducted on four Nvidia A100 GPUs and took about 9 hours per model.

\section{Results}

\subsection{Metrics scores for attribution maps based on bounding boxes or similarity maps} \label{abcd}

The FunnyBirds metrics design follows the Co-12 taxonomy~\cite{nauta2023co}, exploring categories such as Completeness, Correctness, and Contrastivity. The latter two are measured by Single Deletion (SD) and Target Sensitivity (TS) metrics, respectively. At the same time, the Completeness score is calculated as an average of Controlled Synthetic Data Check (CSDC), Preservation Check (PD), Deletion Check (DC), and Distractibility (D).

\begin{figure}[t]
    \centering
    \begin{tabular}{lcccccc}
        & \multicolumn{3}{c}{
            \begin{subfigure}[b]{0.21\textwidth}
                \centering
                \includegraphics[width=\textwidth]{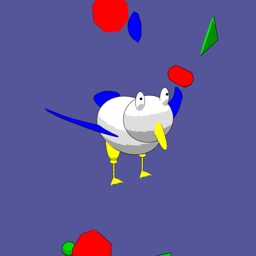}
            \end{subfigure}
        }
        & \multicolumn{3}{c}{
        \begin{subfigure}[b]{0.21\textwidth}
            \centering
            \includegraphics[width=\textwidth]{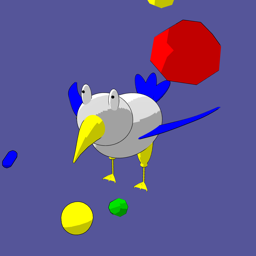}
        \end{subfigure}
        } \\
        & \multicolumn{3}{c|}{
            \begin{subfigure}[b]{0.21\textwidth}
                \centering
                \includegraphics[width=\textwidth]{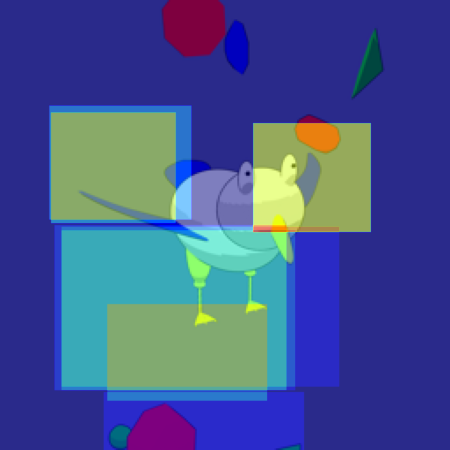}
            \end{subfigure}
            \begin{subfigure}[b]{0.21\textwidth}
                \centering
                \includegraphics[width=\textwidth]{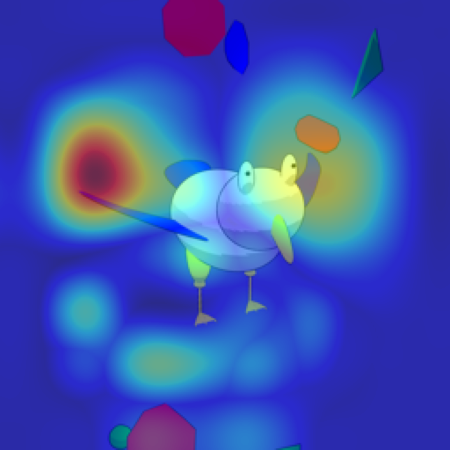}
            \end{subfigure}
        }          
        & \multicolumn{3}{c}{
            \begin{subfigure}[b]{0.21\textwidth}
                \centering
                \includegraphics[width=\textwidth]{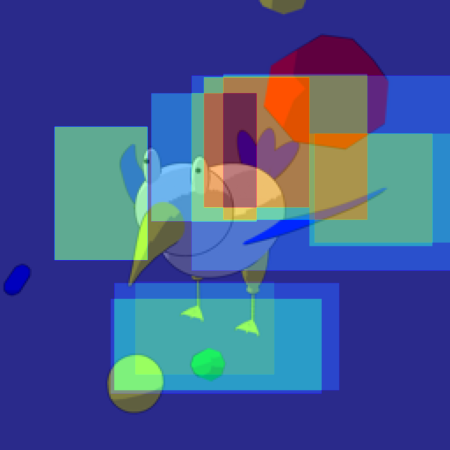}
            \end{subfigure}
            \hfill
            \begin{subfigure}[b]{0.21\textwidth}
                \centering
                \includegraphics[width=\textwidth]{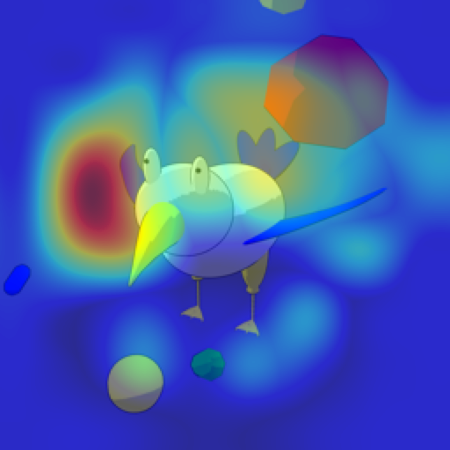}
            \end{subfigure}
        } \\ \hline
        \multicolumn{1}{l|}{Bird part} & \hspace{0.25cm} GT \hspace{0.45cm} & BB &  \multicolumn{1}{c|}{ SSM \hspace{0.25cm}}   & \hspace{0.35cm} GT \hspace{0.45cm} & BB & SSM\hspace{0.25cm} \\ \hline
        \multicolumn{1}{l|}{Foot}     & 1 & 4    & \multicolumn{1}{c|}{3}    & 1     & 5     & 3    \\
        \multicolumn{1}{l|}{Beak}     & 2 & 2    & \multicolumn{1}{c|}{2}    & 4     & 4     & 5    \\
        \multicolumn{1}{l|}{Eye}      & 3 & 3    & \multicolumn{1}{c|}{1}    & 5     & 2     & 4    \\
        \multicolumn{1}{l|}{Wing}     & 4 & 5    & \multicolumn{1}{c|}{5}    & 2     & 2     & 2    \\
        \multicolumn{1}{l|}{Tail}     & 5 & 1    & \multicolumn{1}{c|}{4}    & 3     & 1     & 1    \\ \hline
        \multicolumn{1}{l|}{SD}       &   & 0.35 & \multicolumn{1}{c|}{0.75} &       & 0.40  & 0.75
    \end{tabular}
    \caption{Two sample images (top part), their attribution maps generated based on bounding boxes (BB) or similarity maps (SSM), and corresponding SD scores. As defined in~\ref{eq:sd}, SD is computed as correlation between orders of $\text{PI}(e)$ (GT) and $f(x) - \{f(x_{\setminus p})\}_{p}$ (BB or SSM). We observe that a more precise SSM attribution map demonstrates that ProtoPNet is much more correct than reported in~\cite{hesse2023funnybirds}.}
    \label{fig:SD_examples}
\end{figure}

As presented in Table~\ref{tab:main}, we observe a notable enhancement in explanation correctness as the Single Deletion (SD) score increases from $0.24$ to $0.73$. As defined in~\ref{eq:sd}, SD is computed as correlation between orders of $\text{PI}(e)$ (GT) and $f(x) - \{f(x_{\setminus p})\}_{p}$ (BB or SSM). Therefore, a more precise SSM attribution map demonstrates that ProtoPNet is much more correct than reported in~\cite{hesse2023funnybirds}. We explain this observation using examples in Figure~\ref{fig:SD_examples}.
Moreover, a small increase is observed in its contrastivity, from $0.46$ to $0.5$. 

\begin{table}[t]
    \centering
    \caption{Metric scores obtained for two types of ProtoPNet with ResNet50 visualizations: bounding boxes (BB) and similarity maps (SSM). For SSM, we observe a notable enhancement in correctness and contrastivity but a drop in completeness. This is expected behavior for more precise explanations.}
    \begin{tabular}{llllclc}
        \hline
        \textbf{Co-12 category} & \hspace{.25cm} & \textbf{Metric} & \hspace{.5cm} & \textbf{BB (original)} & \hspace{.5cm} & \multicolumn{1}{l}{\textbf{SSM (ours)}} \\ \hline
                      &  & Accuracy &  & 0.94 &  & 0.93$\pm$0.03 \\
                      &  & BI       &  & 1.00 &  & 1.00$\pm$0.00 \\ \cline{1-3}
        Completeness  &  & CSDC     &  & \textbf{0.93} &  & 0.58$\pm$0.15 \\
                      &  & PC       &  & \textbf{0.91} &  & 0.40$\pm$0.13 \\
                      &  & DC       &  & \textbf{0.92} &  & 0.66$\pm$0.17 \\
                      &  & D        &  & 0.58 &  & \textbf{0.83$\pm$0.04} \\ \cline{1-3}
        Correctness   &  & SD       &  & 0.24 &  & \textbf{0.73$\pm$0.01} \\ \cline{1-3}
        Contrastivity &  & TS       &  & 0.46 &  & \textbf{0.50$\pm$0.07} \\ \hline
    \end{tabular}
    \label{tab:main}
\end{table}

\begin{figure}[t]
    \centering
    \begin{tabular}{ccc|cc}
        &
        \multicolumn{2}{c}{
        \begin{subfigure}[b]{0.21\textwidth}
            \centering
            \includegraphics[width=\textwidth]{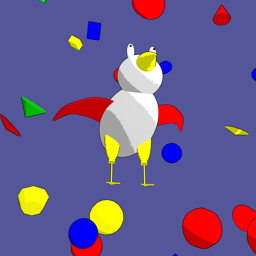}
        \end{subfigure}} &
        \multicolumn{2}{c}{
        \begin{subfigure}[b]{0.21\textwidth}
            \centering
            \includegraphics[width=\textwidth]{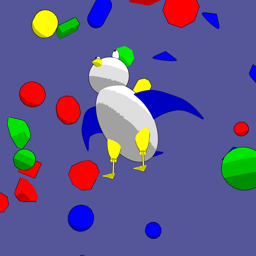}
        \end{subfigure}} \\
        &
        \multicolumn{2}{c|}{
        \begin{subfigure}[b]{0.21\textwidth}
            \centering
            \includegraphics[width=\textwidth]{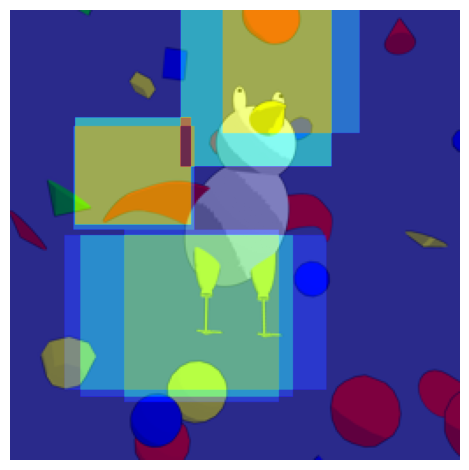}
        \end{subfigure}
        
        \begin{subfigure}[b]{0.21\textwidth}
            \centering
            \includegraphics[width=\textwidth]{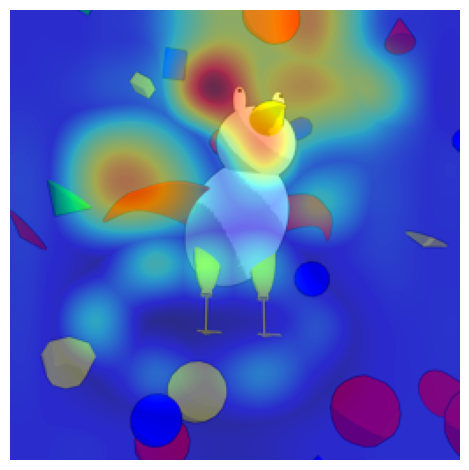}
        \end{subfigure}
        } &
        \multicolumn{2}{c}{
        \begin{subfigure}[b]{0.21\textwidth}
            \centering
            \includegraphics[width=\textwidth]{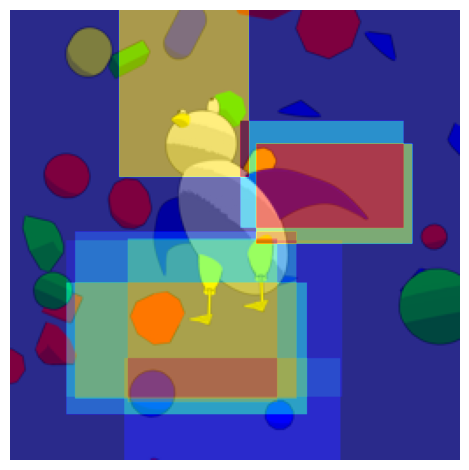}
        \end{subfigure}
        
        \begin{subfigure}[b]{0.21\textwidth}
            \centering
            \includegraphics[width=\textwidth]{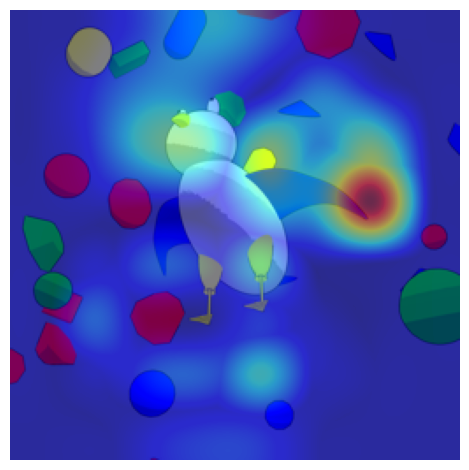}
        \end{subfigure}
        } \\\hline
        \multicolumn{1}{c|}{$t$} &
            \multicolumn{1}{c}{BB} &
            \multicolumn{1}{c|}{SSM} &
            \multicolumn{1}{c}{BB} &
            \multicolumn{1}{c}{SSM} \\ \hline
        \multicolumn{1}{c|}{0.01} &
            \scalebox{.7}{{[}'eye', 'beak', 'foot', 'wing'{]}} &
            \scalebox{.7}{{[}'eye', 'beak', 'foot', 'wing'{]}} &
            \scalebox{.7}{{[}'eye', 'beak', 'foot', 'wing', 'tail'{]}} &
            \scalebox{.7}{{[}'foot', 'wing'{]}} \\
        \multicolumn{1}{c|}{0.1} &
            \scalebox{.7}{{[}'eye', 'beak', 'foot'{]}} &
            \scalebox{.7}{{[}'beak', 'foot', 'wing'{]}} &
            \scalebox{.7}{{[}'eye', 'beak', 'foot', 'wing', 'tail'{]}} &
            \scalebox{.7}{{[}'foot', 'wing'{]}} \\
        \multicolumn{1}{c|}{0.25} &
            \scalebox{.7}{{[}'eye', 'beak', 'foot'{]}} &
            \scalebox{.7}{{[}'wing'{]}} &
            \scalebox{.7}{{[}'eye', 'beak', 'foot', 'wing', 'tail'{]}} &
            \scalebox{.7}{{[}'wing'{]}} \\
        \multicolumn{1}{c|}{0.5} &
            \scalebox{.7}{{[}'eye', 'beak', 'foot'{]}} &
            \scalebox{.7}{{[ }{ ]}} &
            \scalebox{.7}{{[}'eye', 'beak', 'foot', 'wing', 'tail'{]}} &
            \scalebox{.7}{{[ }{ ]}} \\ \hline
        \multicolumn{1}{c|}{\begin{tabular}[c]{@{}l@{}}GT\end{tabular}} &
            \multicolumn{2}{c|}{\scalebox{.7}{\begin{tabular}[c]{@{}c@{}}{[}{[}'beak', 'eye', 'tail'{]}, {[}'beak', 'foot', 'tail'{]},\\ {[}'beak', 'foot', 'wing'{]}, {[}'beak', 'tail', 'wing'{]}{]}\end{tabular}}} &
            \multicolumn{2}{c}{\scalebox{.7}{{[}{[}'tail', 'wing'{]}{]}}}
    \end{tabular}
    \caption{Two sample images (top part), their attribution maps generated based on bounding boxes (BB) or similarity maps (SSM), and important parts ($P$) obtained for various values of $t$. We observe that the original BB approach tends to overidentify parts as important, which results in an incorrectly high completeness score. In contrast, our SSM alternative generates more reliable $P$. Notice that the GT row corresponds to the sets of truly important parts, and the completeness is high if $P$ is similar to one of those sets.} 
    \label{fig:Completness_example}
\end{figure}

Conversely, we observe a significant drop in three out of four completeness metrics. More precisely, the Controlled Synthetic Data Check (CSDC) drops from $0.93$ to $0.58$, the Preservation Check (PC) from $0.91$ to $0.40$, and the Deletion Check (DC) from $0.92$ to $0.66$. As we present in Figure~\ref{fig:Completness_example}, this drop is caused by the fact that the original BB approach tends to overidentify parts as important, which results in an incorrectly high completeness score. In contrast, our SSM alternative generates more reliable $P$. Surprisingly, the remaining completeness metric, Distractibility (D), increases from $0.58$ to $0.83$. This phenomenon may be explained by the fact that D examines irrelevant parts while the remaining metrics concentrate on relevant ones.

These findings underscore the crucial role of visualization techniques within the FunnyBirds framework, particularly in ensuring consistency with other approaches.

\subsection{Various backbones of ProtoPNet}

\begin{figure}[t]
    \centering
    \subfloat[][Input]{
        \label{4figs-a}
        \includegraphics[width=0.20\textwidth]{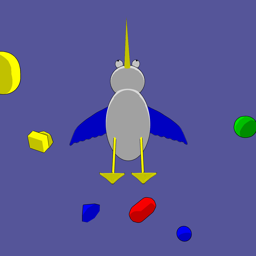}
    }
    \hfill
    \subfloat[][ResNet50]{
        \label{4figs-b} 
        \includegraphics[width=0.20\textwidth]{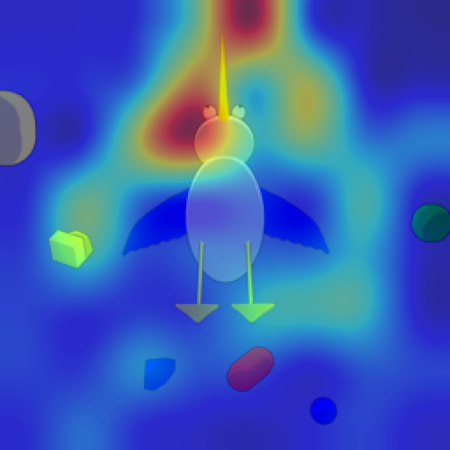}
        }
    \hfill
    \subfloat[][VGG19]{
        \label{4figs-c} 
        \includegraphics[width=0.20\textwidth]{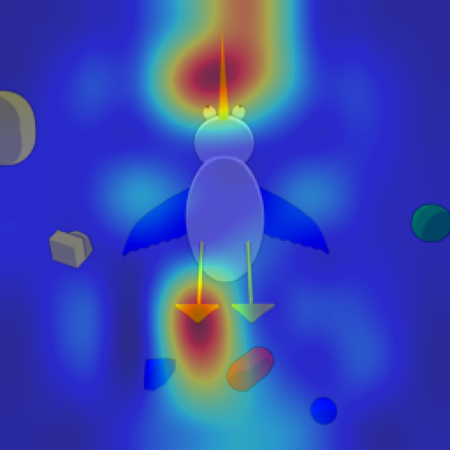}
    }
    \hfill
    \subfloat[][DenseNet169]{
        \label{4figs-d} 
        \includegraphics[width=0.20\textwidth]{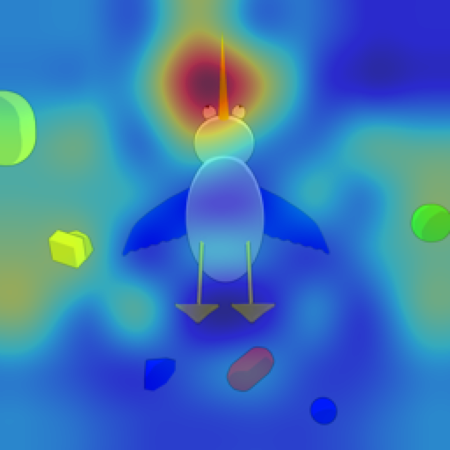}
    }
    \caption{Sample image (a) and SSMs obtained for ProtoPNet with various backbones (b-d).}
    \label{fig:backbone_explanations}
\end{figure}

Table~\ref{tab:backbones} presents metrics scores depending on different backbone architectures (ResNet50, VGG19, and DenseNet169) used in ProtoPNet, while Figure~\ref{fig:backbone_explanations} presents SSM obtained for them. Notably, ResNet50 exhibits substantially higher DC, D, and SD metrics than others, while its TS metric is the lowest. Conversely, for VGG19, CSDC and TS metrics demonstrate superiority. This discrepancy can be attributed to differences in receptive field sizes, notably smaller in the case of VGG19, and the incorporation of bottlenecks in ResNet50.
However, it is important to note that while ResNet50 achieves the best metrics within the FunnyBirds framework, it is outperformed by DenseNet in terms of accuracy.

\begin{table}[t]
    \centering    
    \caption{Metrics scores depending on different backbone architectures (ResNet50, VGG19, and DenseNet169) reveal notable differences. Specifically, ResNet obtains higher explanation metric scores but lower accuracy. It shows the tradeoff between interpretability and accuracy in ProtoPNet.}
    \begin{tabular}{llllclclc}
    \hline
        \textbf{Co-12 category} & \hspace{.25cm} & \textbf{Metric} & \hspace{.5cm} & \textbf{ResNet50} & & \textbf{VGG19} & & \textbf{DenseNet169} \\ \hline
                      &  & Accuracy        &  & 0.93$\pm$0.03          &  & 0.96$\pm$0.00 &  & 0.97$\pm$0.01          \\
                      &  & BI              &  & 1.00$\pm$0.00          &  & 0.99$\pm$0.01 &  & 0.98$\pm$0.01          \\ \cline{1-3}
        Completeness  &  & CSDC            &  & 0.58$\pm$0.15          &  & \textbf{0.63$\pm$0.06} &  & 0.58$\pm$0.04          \\
                      &  & PC              &  & 0.40$\pm$0.13          &  & \textbf{0.48$\pm$0.07} &  & 0.36$\pm$0.06          \\
                      &  & DC              &  & \textbf{0.66$\pm$0.17}          &  & 0.61$\pm$0.10 &  & 0.49$\pm$0.10 \\
                      &  & D               &  & \textbf{0.83$\pm$0.04}          &  & 0.76$\pm$0.01 &  & 0.79$\pm$0.01          \\ \cline{1-3}
        Correctness   &  & SD              &  & \textbf{0.73$\pm$0.01} &  & 0.48$\pm$0.09 &  & 0.49$\pm$0.03          \\ \cline{1-3}
        Contrastivity &  & TS              &  & 0.50$\pm$0.07          &  & \textbf{0.80$\pm$0.05} &  & 0.67$\pm$0.04          \\ \hline
    \end{tabular}
    \label{tab:backbones}
\end{table}

\section{Conclusions}
        
In this study, we evaluated ProtoPNet explanations based on similarity maps rather than bounding boxes within the FunnyBirds framework and comprehensively analyzed the resulting changes in explanation quality. Overall, the results indicate that the choice between bounding boxes and similarity maps significantly impacts the assessment of explanation quality, particularly for methods like ProtoPNet. While bounding boxes have been traditionally used for their simplicity, our study demonstrates that similarity maps provide a more faithful representation of the underlying model's behavior, leading to more accurate evaluation metrics.

Furthermore, our investigation into different backbone architectures highlights the trade-off between interpretability and accuracy inherent in models like ProtoPNet. While models with higher interpretability, such as those based on ResNet50, may achieve lower accuracy, they offer more reliable explanations, as evidenced by higher metric scores. Conversely, models with higher accuracy, such as those based on DenseNet169, may sacrifice interpretability to some extent, resulting in slightly lower metric scores.

In conclusion, our study underscores the importance of carefully considering visualization techniques and model architectures in evaluating explainable AI methods like ProtoPNet. By adopting more precise visualization methods and understanding the trade-offs between interpretability and accuracy, researchers and practitioners can make more informed decisions when deploying and evaluating such models in real-world applications.

\paragraph{Limitations.}
While our study utilizes code provided by the authors of the FunnyBirds framework, it is worth noting that the experiments were conducted on models we trained because the repository did not contain the training code while we prepared this work. This discrepancy could lead to slight variations in results due to model differences. Nevertheless, we tried to mitigate this by reporting metrics scores averaged over multiple runs.

\paragraph{Impact.}
This research addresses the challenge of evaluating different eXplainable Artificial Intelligence (XAI) methods, particularly comparing post-hoc and ante-hoc approaches. Leveraging a recently published framework, we emphasize the importance of unifying approaches for deriving metric scores and advocate for using similarity map-based explanations of prototypical parts when evaluating and comparing them with saliency-based methods.

\bibliographystyle{splncs04}
\bibliography{main}

\end{document}